\documentclass[runningheads]{llncs}
\usepackage{graphicx}

\usepackage{tikz}
\usepackage{comment}
\usepackage{amsmath,amssymb} 
\usepackage{color}

\newcommand{\etal}{\textit{et al}.}
\usepackage{multirow}
\usepackage{booktabs}
\usepackage{threeparttable}

\begin{document}
\pagestyle{headings}
\mainmatter
\def\ECCVSubNumber{2720}  

\title{ProbNVS: Fast Novel View Synthesis with Learned Probability-Guided Sampling} 

\titlerunning{ProbNVS}
%
\author{Yuemei Zhou\inst{1} \and
Tao Yu\inst{1} \and
Zerong Zheng\inst{1} \and
Ying Fu\inst{2} \and
Yebin Liu\inst{1}}
%
\authorrunning{Y. Zhou et al.}
%
\institute{Department of Automation, Tsinghua University \and 
Beijing Institute of Technology
}
\maketitle

\begin{abstract}
Existing state-of-the-art novel view synthesis methods rely on either fairly accurate 3D geometry estimation or sampling of the entire space for neural volumetric rendering, which limit the overall efficiency. 
In order to improve the rendering efficiency by reducing sampling points without sacrificing rendering quality, we propose to build a novel view synthesis framework based on learned MVS priors that enables general, fast and photo-realistic view synthesis simultaneously. Specifically, fewer but important points are sampled under the guidance of depth probability distributions extracted from the learned MVS architecture. Based on the learned probability-guided sampling, a neural volume rendering module is elaborately devised to fully aggregate source view information as well as the learned scene structures to synthesize photorealistic target view images. Finally, the rendering results in uncertain, occluded and unreferenced regions can be further improved by incorporating a confidence-aware refinement module. Experiments show that our method achieves 15 to 40 times faster rendering compared to state-of-the-art baselines, with strong generalization capacity and comparable high-quality novel view synthesis performance.
\end{abstract}

\section{Introduction}

Novel view synthesis~(NVS) from a set of posed images of a scene is a long-standing problem in computer vision and graphics. Generating photo-realistic novel view images with high rendering speed is essential for immersive experience 
and has broad applications in sports broadcasting, holographic communications, etc. 
The study of novel view synthesis can be traced back to view interpolation by morphing~\cite{chen1993view,mcmillan1995plenoptic}, rendering combined with modeling~\cite{debevec1996modeling}, and light field rendering from dense sampling of scenes~\cite{levoy1996light,gortler1996lumigraph,buehler2001unstructured}. 
With the advent of deep learning era, deep neural networks learned from a large number of posed image sets~\cite{flynn2016deepstereo,kalantari2016learning,zhou2018stereo} were introduced to synthesize novel views and have achieved breakthroughs on generality and image quality. 

Regarding to the popular learning-based NVS, one line of work is ``geometry-based NVS'' that learns to aggregate multi-view information based on 3D geometry estimation~\cite{dai2020neural,thies2019deferred,aliev2020neural,riegler2020free,riegler2020stable}, and another line is ``radiance field-based NVS'' that represents the scenes as neural radiance fields which optimize density and color fields implicitly~\cite{yu2021pixelnerf,martin2021nerf,chibane2021stereo,chen2021mvsnerf,wang2021ibrnet}.
The geometry-based methods heavily rely on accurate scene geometry initialization. A 3D geometry scaffold is firstly estimated with the help of traditional Structure-from-Motion~(SfM)~\cite{schoenberger2016sfm}, Multi-view Stereo~(MVS)~\cite{schoenberger2016mvs} and meshing, and then pixels on novel views are rendered by aggregating multi-view features accordingly. The pre-calculation of scene geometry is very time-consuming, which usually takes tens of minutes for each new scene. Besides, this type of methods premises accurate scene geometry estimation, and the geometry errors will be immediately reflected on the synthesised image. 
In contrast, the radiance field-based methods extend the popular and extensively studied Neural Radiance Field~(NeRF)~\cite{mildenhall2020nerf} to have generalization abilities by conditioning on various additional factors, such as image features~\cite{yu2021pixelnerf,wang2021ibrnet}, stereo correspondences~\cite{chibane2021stereo} and plane-swept cost volumes~\cite{chen2021mvsnerf}.  However, due to the need of querying a large number of 3D points scattered throughout the space, these methods generally take 
much longer time than geometry-based methods in rendering a novel view image.
Although methods to further accelerate the rendering of neural radiation fields have been proposed, they all focus on optimizing a scene-specific model. They either divide the space into a sparse voxel octree or thousands of tiny MLPs~\cite{liu2020neural,yu2021plenoctrees}, or pre-sample NeRF into a tabulated view-dependent volume and render from the cache~\cite{garbin2021fastnerf,reiser2021kilonerf}.

Our goal is to achieve efficient and general novel view rendering by reducing the number of sampling points while maintaining good quality of radiance field rendering. 
To fulfill this goal, we present ProbNVS, a novel view synthesis framework that enables fast, generic and photo-realistic view synthesis simultaneously. The key idea of our work is leveraging learned depth probability distributions to guide the sampling for neural volume rendering.
To this end, we first present multi-scale probability-guided sampling. Specifically, we construct multi-scale cost volumes on the target view frustum, and let the network learn the corresponding depth probability distributions to guide the sampling in a coarse-to-fine manner. The finest set of sampling points, or the ``points for rendering'', along with their depth probabilities are utilized for later neural volume rendering. The probability-guided sampling not only preserves the realistic rendering quality of neural volume rendering, but also significantly improves the rendering efficiency with much fewer sampling points. 
Secondly, a neural volume rendering module is elaborately designed to fully aggregate source view information as well as the learned scene structures to synthesize target view images. At last, a confidence-aware refinement module is presented to further improve the rendering results of uncertain, occluded and unreferenced areas, where the confidence is obtained by the depth probabilities learned from the probability-guided sampling module. 



Above all, our ProbNVS is a general NVS architecture, and the learned probability-guided sampling can significantly reduce the sampling points, thereby dramatically speeding up the rendering procedure. Additionally, the accuracy and realism of the rendered image, especially the performance on scenes with hidden components, occluded regions and non-Lambertian surfaces, can be promoted by the subsequent neural volume rendering and confidence-aware refinement. 
Moreover, our framework is fully differentiable and can generalize across scenes without re-training for new scene adaption. Experiments demonstrate that our ProbNVS shows accurate and realistic photorealistic novel view synthesis ability across various scenes. Our ProbNVS achieves an outstanding rendering speed of around 0.15s for synthesising a 640$\times$480 image, which is about 15$\times$, 38$\times$ and 338$\times$ faster than MVSNeRF~\cite{chen2021mvsnerf}, IBRNet~\cite{wang2021ibrnet} and pixelNeRF~\cite{yu2021pixelnerf}, respectively. In summary, our key contributions are:

\begin{itemize}
\setlength{\itemsep}{0pt}
\setlength{\parsep}{0pt}
\setlength{\parskip}{0pt}
\vspace{-0.2cm}

    \item We propose ProbNVS, a generic novel view synthesis framework that leverages learned MVS priors for fast and photorealistic view synthesis.
    \item We design a probability-guided sampling strategy, that multi-scale cost volumes are constructed on the target view frustum, and the network learns to sampling points according to the depth probability distributions in a coarse-to-fine manner. The learned probability-guided sampling significantly speeds up the rendering without losing realistic rendering capacity.
    (Sect.~\ref{sec:sampling})
    \item We introduce a subsequent neural volume rendering module~(Sect.~\ref{sec:construction}) and a confidence-aware refinement module~(Sect.~\ref{sec:refine}) for photorealistic novel view synthesis even in the occluded and boundary regions. 
    
\vspace{-0.2cm}
\end{itemize}

\section{Related Work}


Novel view synthesis has been an appealing research problem in computer vision and graphics since the 1990s. The early work~\cite{chen1993view} by Chens and Williams opened up a new research topic for image synthesis via view interpolation, where the 3D scene representation can be replaced by posed images. 
Later on, Debevec \etal~\cite{debevec1996modeling} proposed a hybrid approach composed of both geometry modeling and image warping from a set of still photographs. In the meantime, light field rendering~\cite{levoy1996light,gortler1996lumigraph} was proposed as a research branch by constructing a 4D light field and interpolating views without image warping or geometry information.

A key issue in NVS is how to model the scene from only 2D captures.
Some work avoids modeling the scene geometry and implements NVS through view morphing~\cite{chen1993view,seitz1996view,fitzgibbon2005image} by utilizing pixel correspondences and projective mapping, or light field rendering~\cite{levoy1996light,levin2010linear,shi2014light,vagharshakyan2015image,vagharshakyan2017light,wu2017light,yoon2015learning,wang2017light,wu2018light,wu2019learning}. However, to generate 3D realistic novel views without dense view capture like light fields, modeling the underlying geometry information of scenes plays an important role.
Therefore, we further divide the previous work into explicit and implicit modeling of scene structures: novel view synthesis~(or image-based rendering) (1) with explicit geometry estimation and (2) with implicit scene representation.


\noindent\textbf{NVS with explicit geometry estimation.}
To generate realistic and immersive novel views, some work focused on rendering based on explicit geometric estimation of pose images.
Chaurasia \etal~\cite{chen1993view} introduced superpixels augmented with synthesized depth and generated novel views by blending the warped superpixels from adjacent input views. Zitnick \etal~\cite{zitnick2004high} developed a two-layer representation with matting for video view interpolation. With the prospering of deep learning, Flynn \etal~\cite{flynn2016deepstereo} presented an end-to-end deep architecture that learns depth and color rendering together. Penner \etal~\cite{penner2017soft} presented a soft 3D reconstruction pipeline that preserves depth uncertainty throughout the geometry estimation and rendering pipeline. Choi \etal~\cite{choi2019extreme} incorporated the concept of depth probability volumes as the representation of scene geometry to guide view extrapolation. Given a pre-computed 3D geometry scaffold, a line of work~\cite{riegler2020free,riegler2020stable} focused on neural rendering in terms of the reconstructed surface of the scene, enabling sharp and realistic novel view synthesis on the condition of complete and accurate geometry estimation. These methods either requires complex and time-consuming preprocessing or relies on fairly accurate depth estimates.

\noindent\textbf{NVS with implicit scene representations.}
Instead of learning explicit scene geometry and color rendering separately, Zhou \etal~\cite{zhou2018stereo} proposed a multiplane image representation that represented the target scene with a set of parallel RGB-alpha layers inspired by layered depth image~\cite{shade1998layered}. It allows end-to-end training with posed stereo images and supports differentiable view interpolation and extrapolation. Following this trend, plenty of view synthesis methods based on this scene representation~\cite{srinivasan2019pushing,mildenhall2019local,flynn2019deepview,tucker2020single} have been proposed to render more photorealistic novel view images for scenes collected on a plane. In the last few years, representing scenes implicitly with neural networks for 3D reconstruction~\cite{park2019deepsdf,mescheder2019occupancy,peng2020convolutional,jiang2020local,genova2020local} has been a trending research topic, which leverages MLPs to map continuous spatial coordinates to signed distance or occupancy indicating the surface locations. Later, Mildenhall \etal~\cite{mildenhall2020nerf} proposed Neural Radiance Field~(NeRF) that achieved impressive view synthesis results by optimizing a 5D continuous radiance field for each scene. The proposal of this method has inspired many related researches~\cite{zhang2020nerf++,wang2021nerf,martin2021nerf} that further improved the performance of neural rendering. Some work were proposed to accelerate rendering process by dividing the space into a sparse voxel octree~\cite{liu2020neural}, leveraging thousands of tiny MLPs~\cite{reiser2021kilonerf}, pre-sampling NeRF into a tabulated view-dependent volume or rendering from the cache~\cite{yu2021plenoctrees,garbin2021fastnerf}.
Some other work endowed neural volume rendering with generalization capabilities by introducing pixel features~\cite{yu2021pixelnerf} or correspondence matching~\cite{chibane2021stereo,wang2021ibrnet,chen2021mvsnerf}. 
 Although high-speed rendering is achieved by those methods, they are mainly scene-specific and cannot be generalized to new scenes. By contrast, the existing general neural rendering methods suffer from long processing time, e.g., rendering a 640$\times$512 image takes at least a few seconds.
 Therefore, to achieve photorealistic novel view synthesis with high rendering speed and generalization abilities, we propose a general NVS framework with a novel depth probability-guided sampling strategy. By incorporating learned probability-guided sampling, the number of sampling points can be significantly reduced, resulting in 15 to 40 times faster rendering.

\begin{figure}[t]
\begin{center}
\includegraphics[width=0.98\linewidth]{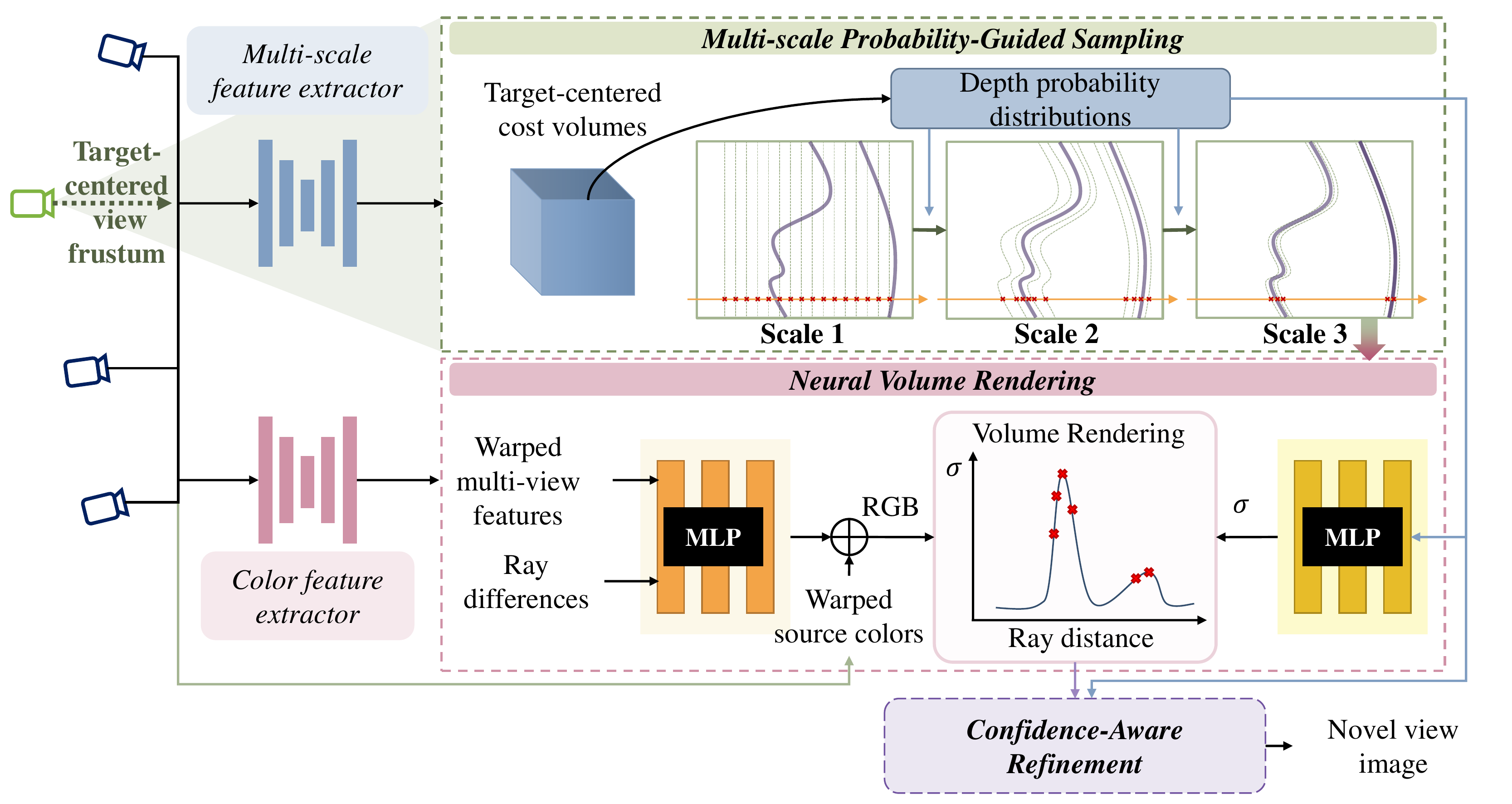}
\end{center}
\vspace{-5mm}
\caption{Method overview. Our ProbNVS is composed of three main parts: multi-scale probability-guided sampling~(light green box), neural volume rendering~(light pink box) and confidence-aware refinement~(light purple box). The probability-guided sampling module performs coarse-to-fine sampling based the learned depth probability distributions from the target-centered cost volumes, and the finest samples are feed into the neural volume rendering module for novel view synthesis. A confidence-aware refinement module is proposed to improve the results of uncertain, occluded and unreferenced areas when the input source views are limited.}
\vspace{-5mm}
\label{fig:pipeline}
\end{figure}

\section{Method}

Given a set of posed images of a real scene, our goal is to 
synthesize photorealistic novel view images in the blink of an eye without the need for exhaustive pre-calculation of the scene geometry or sampling throughout the entire target space.
Specifically, the pipeline (Fig.~\ref{fig:pipeline}) is composed of three main parts: (1) the multi-scale probability-guided sampling where sampling is guided by learned depth probability distributions in a coarse-to-fine manner~(Sect.~\ref{sec:sampling}), 
(2) a neural volume rendering module that generates the density and color fields of the rendering points through aggregating source view information and the learned depth probabilities~(Sect.~\ref{sec:construction})
, and (3) a confidence-aware refinement module that further improves rendering quality especially for the uncertain, occluded and unreferenced regions~(Sect.~\ref{sec:refine}). 

\begin{figure}[t]
\begin{center}
\includegraphics[width=0.98\linewidth]{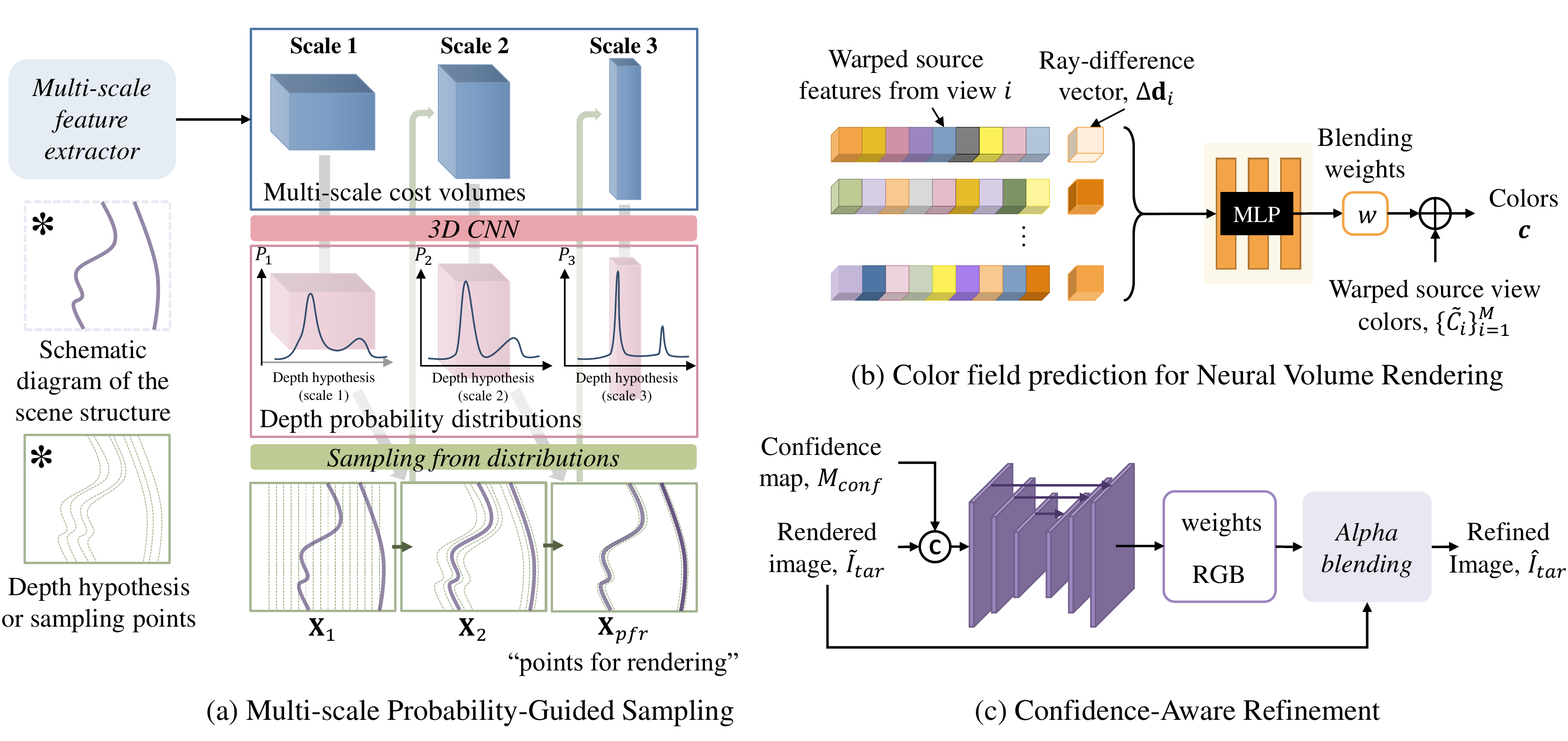}
\end{center}
\vspace{-5mm}
\caption{Details of the ProbNVS sub-modules. In (a) multi-scale probability-guided sampling, we first constructed target-centered cost volumes from the variance of the source view features, and regularize the cost volumes with 3D CNN to learn depth probability distributions. Then points can be sampled guided by the learned depth probability distributions in a coarse-to-fine manner. Based on the finest ``points for rendering'', the (b) color field of neural volume rendering is estimated by aggregating source view features, ray-difference vectors and source view colors. The (c) confidence-aware refinement module is conditioned on the confidence calculated from the learned depth probability distributions, and the refined image is generated by alpha compositing.}
\vspace{-5mm}
\label{fig:method}
\end{figure}

\subsection{Multi-scale Probability-Guided Sampling}
\label{sec:sampling}

In this section, we present a multi-scale probability-guided sampling strategy, which samples points that contribute significantly to view synthesis in a coarse-to-fine manner guided by the learned depth probability distributions.
The input of the multi-scale probability-guided sampling is the target camera pose $\Phi_{tar}$ and the nearest $M$ source view images $\{I_i\}_{i=1}^M$ with their camera poses $\{\Phi_i\}_{i=1}^M$. The output is a set of ``points for rendering'' $\mathbf{X}_{pfr}\in \mathbb{R}^{D_3 \times H\times W}$ on the target view frustum, where $H\times W$ is the spatial resolution of the input images and $D_3$ is the number of sampling points along each ray from the finest sampling level.

\noindent\textbf{Multi-scale feature extractor.} 
Given the selected input views $\{I_i\}_{i=1}^M$, we first extract multi-scale features $\{F_{i,1}, F_{i,2}, F_{i,3}\}_{i=1}^M$ with a multi-scale feature extractor, implemented as a 2D UNet with instance normalization (IN)~\cite{ulyanov2016instance}.
The extracted multi-scale features $F_{i,1}, F_{i,2}, F_{i,3}$ have spatial resolution of $\frac{H}{4}\times \frac{W}{4}, \frac{H}{2}\times \frac{W}{2}, H\times W$ and channel numbers of $C_1, C_2, C_3$ respectively.

\noindent\textbf{Target-centered cost volumes.}
In order to learn the 3D structural information of the target scene for guiding sampling, we build cost volumes on the target view frustum. 
The target-centered cost volumes are constructed on the target view frustum via differentiable reprojection of the extracted source view features. This is slightly different from accustomed learning based MVS methods~\cite{yao2018mvsnet,cheng2020deep} where plane-swepted cost volumes are built on views with known images. Specifically, given the target camera pose $\Phi_{tar} = \{\mathbf{K}_{tar}, \mathbf{R}_{tar}, \mathbf{t}_{tar}\}$, the source view poses $\{\Phi_i = \{\mathbf{K}_i, \mathbf{R}_i, \mathbf{t}_i\}\}_{i=1}^M$ and a sampling point on the target view frustum with depth $d$, we consider the warping matrix $\mathbf{H}_i(d)$ that warps features from the source view coordinate to the target view coordinate:

\begin{equation}
\begin{aligned}
\mathbf{H}_i(d) &= \hat{\mathbf{K}}_i \mathbf{T}_i \mathbf{T}_{tar}^{-1}\hat{\mathbf{K}}_{tar}^{-1}, \\
\mathbf{x}_i &= \mathbf{H}_i(d)\mathbf{x}_{tar}.
\label{eq:warping}
\end{aligned}
\end{equation}
where $\hat{\mathbf{K}} = \left( \begin{array}{cc} \mathbf{K} & \mathbf{0} \\ \mathbf{0}^T & 1 \end{array} \right) \in \mathbb{R}^{4\times4}$, $\hat{T} = \left( \begin{array}{cc} \mathbf{R} & \mathbf{t} \\ \mathbf{0}^T & 1 \end{array} \right) \in \mathbb{R}^{4\times4}$, and $\mathbf{x}_i$ and $\mathbf{x}_{tar}$ are the homogeneous vector $(d_i u_i, d_i v_i, d_i, 1)^T$ and $(d u_{tar}, d v_{tar}, d, 1)^T$, corresponding to the pixel coordinates $(u_i, v_i)^T$ and $(u_{tar}, v_{tar})^T$ at the source view and the target view respectively. Given a set of sampling points $\mathbf{X}\in \mathbb{R}^{D \times H\times W}$ on the target view frustum, a target-centered cost volume is constructed by the variance of the warped features across views.
We then regularize the cost volume with a 3D CNN.
After applying the \textit{softmax} operation to the regularized cost volume along the depth direction for probability normalization, we obtain the corresponding depth probability distributions $P \in \mathbb{R}^{D \times H\times W}$ of the given set of samples $\mathbf{X}$. Most importantly, the learned depth probability distributions give hints on how to guide the sampling for a more compact solution space.

\noindent\textbf{Multi-scale depth probability-guided sampling.}
In order to speed up the rendering by reducing the sampling points without losing realistic rendering capacity, we present multi-scale depth probability-guided sampling. The sampling procedure is performed in a coarse-to-fine manner at three different scales, where points that contribute more to rendering can be gradually sampled guided by the learned depth probability distributions.
Different from the single peak assumption in~\cite{cheng2020deep}, we adopt a strategy of sampling from distribution, which takes into account the probability of sampling along a entire ray.

In detail, we devise a coarse-to-fine sampling strategy in three scales which are commonly used in learning-based MVS architectures~\cite{yao2018mvsnet,yao2019recurrent,gu2020cascade,cheng2020deep}, and each following scale constructs a target-centered cost volume to refine the previous sampling with higher pixel resolution and finer probability-guided sampling. In the coarsest scale, points are uniformly sampled along camera rays on the target view frustum within a predefined depth interval $(d_{min}, d_{max})$, denoted as $\mathbf{X}_1 \in \mathbb{R}^{D_1 \times \frac{H}{4}\times \frac{W}{4}}$. Subsequently, a target-centered cost volume can be constructed from the image features extracted at the first scale $\{F_{i,1}\}_{i=1}^M$. After regularization with a 3D CNN and normalization, the depth probabilities of the coarsest samples $P_1 \in \mathbb{R}^{D_1 \times \frac{H}{4}\times \frac{W}{4}}$ can be learned accordingly. Afterwards, inverse transform sampling is used to sample a smaller set of points $\mathbf{X}_2' \in \mathbb{R}^{ D_2\times\frac{H}{4}\times \frac{W}{4}}$ with more possibilities to represent scene structure according to $P_1$. We then interpolate spatial resolution of the samples to the next scale $\mathbf{X}_2 \in \mathbb{R}^{D_2\times\frac{H}{2}\times \frac{W}{2}}$. In the second scale, we follow the same procedure of cost volume construction, regularization and normalization, and obtained the depth probability distributions of this scale $P_2 \in \mathbb{R}^{D_2\times \frac{H}{2}\times \frac{W}{2}}$. Then, we continue to sample a finer set of samples as ``points for rendering'' $\mathbf{X}_{pfr} \in \mathbb{R}^{D_3\times H \times W}$ according to the $P_2$. Similarly, we can obtain corresponding depth probability distributions $P_3 \in \mathbb{R}^{D_3\times H \times W}$. In brief, embedding the depth probability-guided sampling into a multi-scale MVS learning procedure enables more compact and efficient sampling, which can significantly speed up rendering without losing realistic rendering capacity.


\subsection{Neural Volume Rendering}
\label{sec:construction}
In this section, we present a neural volume rendering module that takes the ``points for rendering'' and the corresponding depth probabilities from Sect.~\ref{sec:sampling} as inputs and outputs photorealistic novel view images. To perform neural volume rendering as in NeRF~\cite{mildenhall2020nerf}, we represent the scene observed by the target viewpoint as density and color fields, which are built based on the delicate sampling points at the target view by fully aggregating the source view information as well as the learned scene structures.

\noindent\textbf{Density prediction.}
The densities encode the transmittance of the 3D components in the target space. In most cases, the 3D space of the target view frustum is mostly empty with finite objects. Therefore, we utilize the finest rendering points from Sect.~\ref{sec:sampling} and estimate their densities to fit the real scene structures. Moreover, the leaned depth probabilities of the rendering points imply the 3D structure of the target scene. As a result, taking the learned depth probabilities of the ``points for rendering'' $\mathbf{X}_{pfr}$ as inputs, we introduce a density estimation MLP $f_{\sigma}$ to predict the densities of corresponding samples.
\begin{equation}
\begin{aligned}
\sigma(r) = f_{\sigma}(P_3(r)),
\label{eq:mlp_density}
\end{aligned}
\end{equation}
where $P_3(r) \in \mathbb{R}^{D_3}$ is the learned depth probability distribution of a target camera ray $r$,
and $\sigma(r) \in \mathbb{R}^{D_3}$ is the densities of sampling points on the target camera ray.
Moreover, gathering the densities $\sigma(r)$ of rendering points gives us the density fields $\mathbf{\zeta} \in \mathbb{R}^{D_3\times H \times W}$.

\noindent\textbf{Color prediction.}
After the density estimation, we predict the surface colors by aggregating the source view color information based on the finest rendering points from Sect.~\ref{sec:sampling}. We find color blending shows more promising results and is easier to learn, so we aggregate multi-view colors by blending them with the estimated blending weights. Specifically, we first calculate the blending weights by thoroughly considering of the extracted features of the source views
and the relative viewing directions. Then, we warp the source view colors using the warping matrix as in Eq.~(\ref{eq:warping}) and colors of the ``points for rendering'' are blended using the learned blending weights. 

Specifically, we first extract another set of image features $\{F_i^c\}_{i=1}^M$ with a 2D U-Net for the color blending weights generation, which have spatial resolution of $H\times W$ and channel number of $C_3$. We then warp the extracted multi-view features to the target view following the warping procedure as in Eq.~(\ref{eq:warping}), and the warped features are denoted as $\{\tilde{F}_i^c\}_{i=1}^M$. After that, we calculate the relative viewing directions $\{\Delta \mathbf{d}_i\}_{i=1}^M$, which is the difference between the directions of the target ray and source rays, i.e., $\Delta \mathbf{d}_i = \mathbf{d} - \mathbf{d}_i$. With the warped image features and the ray direction difference as inputs, blending weights can then be generated by applying an MLP network, i.e., $w_i = f_{color}(\tilde{F}_i^c, \Delta \mathbf{d}_i)$ Finally, the colors of the ``points for rendering'' can be blended using the warped multi-view colors $\{\tilde{C}_i\}_{i=1}^M$ and the blending weights $\{w_i\}_{i=1}^M$ through a \textit{softmax} operator.
\begin{equation}
\begin{aligned}
\mathbf{c} &= \sum_{i=1}^M(\tilde{C}_i \exp{(w_i)}/\sum_{j=1}^M \exp{(w_j)}),
\label{eq:mlp_color}
\end{aligned}
\end{equation}
and gathering $\mathbf{c}$ of the rendering points gives us the color fields $\mathbf{C} \in \mathbb{R}^{3\times D_3\times H \times W}$.

\noindent\textbf{Volume rendering.}
With the learned density and color fields, we leverage the vanilla volume rendering to obtain the synthesised target view image~\cite{mildenhall2020nerf}: 
\begin{equation}
\begin{aligned}
\tilde{C}_r = \sum_{k=1}^{D_3} T_k(1-\exp{(-\sigma_k)})\mathbf{c}_k, \\
\text{where}\quad T_k = \exp{(-\sum_{m=1}^{k-1}\sigma_m)},
\label{eq:volume_rendering}
\end{aligned}
\end{equation}
where $\tilde{C}_r$ is the accumulated pixel color of the predicted target image corresponding to ray $r$, $D_3$ is number of sampled surface points on the ray, and $\{\sigma_k, \mathbf{c}_k\}$ are the density and color of the $k$th sampling point, and $T_k$ represents the transmittance of the $k$th point. 

\subsection{Confidence-aware Refinement}
\label{sec:refine}

After accumulating the colors and densities of the rendering points with volume rendering, we obtain the synthesised target view image $\tilde{I}_{tar}$. However, artifacts may appear in unreferenced, occluded and textureless regions due to the difficulty of correspondence estimation with limited adjacent source views as input~($M=3$ for instance).
Thankfully, we can infer artifacts that might appear on the synthetic target view image from the confidence of probability-guided sampling. As a result, we propose a confidence-aware refinement module, i.e.,
\begin{equation}
\begin{aligned}
\mathbf{\alpha}, \hat{I}_{mid} = f_{u}(\tilde{I}_{tar}, M_{conf}), \\
\hat{I}_{tar} = \mathbf{\alpha}\tilde{I}_{tar} + (1-\mathbf{\alpha})\tilde{I}_{mid},
\label{eq:refine}
\end{aligned}
\end{equation}
where $f_{u}$ is the confidence-aware refinement module which learns to refine the uncertain, unreferenced~(e.g. image boundaries) and occluded area for better visual fidelity, and $M_{conf}\in \mathbb{R}^{H \times W}$ is the confidence map which represents the confidence of our multi-scale probability-guided sampling. We obtain the sampling confidence $M_{conf}$ by summing the interpolated depth probabilities of the ``points for renderng'' corresponding to the normalized depth probability estimated at the first scale. $\tilde{I}_{mid}$ is the intermediate output image, and $\mathbf{\alpha}$ is the learned alpha map. The implementation details and calculation of the sampling confidence will be discussed in the supplemental material.

\subsection{Losses}
\label{sec:loss}
\noindent\textbf{Multi-scale depth loss.}
We adopt the multi-scale depth loss for initializing the multi-scale probability-guided sampling. First, we can obtain depth maps of three scales $\{d_i\}_{i=1}^3$ by multiplying the depth labels and depth probability distributions. Then, we apply $L_1$ loss for three stages.
\begin{equation}
\mathcal{L}_{depth} = \lambda_{1}\lVert{\hat{d}_1-d_1}\rVert_1 + \lambda_{2}\lVert{\hat{d}_2-d_2}\rVert_1 + \lambda_{3}\lVert{\hat{d}_3-d_3}\rVert_1,
\label{loss:depth}
\end{equation}

\noindent\textbf{Rendering loss.}
We adopt $L_1$ loss and perceptual loss for the synthesised novel view image output $\tilde{I}_{tar}$.

\begin{equation}
\mathcal{L}_{render} = \lVert{\tilde{I}_{tar}-I_{tar}}\rVert_1 + \sum_{l}{\lVert{\phi_l(\tilde{I}_{tar})-\phi_l(I_{tar})}\rVert_1},
\label{loss:render}
\end{equation}
where $\{\phi_l\}$ is a set of appointed and pre-trained neural layers of VGG-16.

\noindent\textbf{Refinement loss.} For the refinement loss, we adopt $L_1$ for pixel reconstruction loss.

\begin{equation}
\mathcal{L}_{refine} = \lVert{\phi_l(\hat{I}_{tar})-\phi_l(I_{tar})}\rVert_1 + \lVert{\phi_l(\hat{I}_{mid})-\phi_l(I_{tar})}\rVert_1.
\label{loss:refine}
\end{equation}
where $\hat{I}_{tar}$ is the refined image, and $\hat{I}_{mid}$ is the intermediate output image of the confidence-aware refinement module in Sect.~\ref{sec:refine}.

\subsection{Implementation Details}
\label{sec:implement}


\noindent\textbf{Network details.} The multi-scale feature extractor is vanilla 2D U-Net with instance normalization, and the number of feature channels in three scales are $C_1=32, C_2=16, C_3=8$. The 3D CNNs for regularizing the cost volume of three scales are 3D U-Net with skip connections and instance normalization. The numbers of depth samples in three scales are $D_1=64, D_2=32, D_3=8$ respectively, and we use $D_3=8$ for the neural volume rendering. Please refer to the supplementary material for more details.

\noindent\textbf{Training details.} We train our ProbNVS with $M=3$ nearest source views for NVS with a image resolution of $W\times H=640\times 512$ on DTU dataset~\cite{aanaes2016large}.
We pre-train the multi-scale probability-guided sampling with the multi-scale depth loss first, as a better initialization can accelerate the convergence of the remaining rendering. After that, we train the sampling and neural rendering together on DTU dataset~\cite{aanaes2016large} with the rendering loss in Eq.~(\ref{loss:render}). In order to further improve the results on unreferenced and occluded regions, we freeze the rendering part of network and train the refinement module. The learning rate is 0.0016 for pre-training the probability-guided sampling, and 0.0005 for training the whole rendering pipeline, and 0.001 for the refinement network. We adopt Adam~\cite{kingma2014adam} for all the training procedures. We train the three steps for 60 epochs, 15 epochs and 300k iterations on two GTX 3090 GPUs.

\begin{table}[t]
\begin{center}
\caption{Quantitative results of novel view synthesis on the DTU dataset.}
\label{table:dtu}
\setlength{\tabcolsep}{2mm}
\begin{tabular}{l c c c c c}
\hline
\multicolumn{1}{l}{\multirow{2}{*}{Method}}&
\multicolumn{1}{c}{\multirow{2}{*}{\shortstack{Points for\\rendering}}} &\multicolumn{1}{c}{\multirow{2}{*}{\shortstack{Rendering\\speed/(fps)$\uparrow$}}} & \multicolumn{3}{c}{DTU test set}\\ \cmidrule(r){4-6}
                         & &  & PSNR$\uparrow$ & SSIM$\uparrow$ & LPIPS$\downarrow$ \\ 
\hline
pixelNeRF~\cite{yu2021pixelnerf}                & 64+128 & 0.0174 & 20.3153 & 0.632 & 0.5093\\
IBRNet~\cite{wang2021ibrnet}                    & 64+64 & 0.1567 & 21.5963 & 0.7494 & 0.3957\\
MVSNeRF~\cite{chen2021mvsnerf}                  & 128 & 0.3897 & \underline{24.3726} & \textbf{0.8453} & 0.3269\\
\hline
ProbNVS w/o Refine                     & 8 & \textbf{6.5342} & 23.8093 & 0.8167 & \underline{0.2622}\\
ProbNVS                            & 8 & \underline{5.8824} & \textbf{24.6635} & \underline{0.8316} & \textbf{0.2516}\\
\hline
\end{tabular}
\end{center}
\vspace{-4mm}
\end{table}

\section{Experiments}
\label{sec:exp}
In this section, we evaluate our method on three different datasets~\cite{aanaes2016large,mildenhall2020nerf,mildenhall2019local} from synthetic data to real data~(Sect.~\ref{sec:eval_data}), and give quantitative and qualitative comparisons with state-of-the-art generic novel view synthesis work~(Sect.~\ref{sec:comparison}). Moreover, we conduct an ablation study on our ProbNVS architecture, and then analysize the relationship between NVS quality and the number of rendered samples and input source views~(Sect.~\ref{sec:ablation}). 

\subsection{Datasets}
\label{sec:eval_data}
We first evaluate our model on the DTU dataset~\cite{aanaes2016large} consisting of 124 different scenes with diverse objects, collected under seven different lighting conditions. Each scene has 49 renderings with a spatial resolution of $640\times 512$ pixels. We use the same dataset partition protocol as pixelNeRF~\cite{yu2021pixelnerf} and MVSNeRF~\cite{chen2021mvsnerf} by splitting the dataset into 88 training scenes, 15 validation scenes and 16 testing scenes. To further examine the generalization ability, We evaluate our method on the NeRF synthetic dataset~\cite{mildenhall2020nerf} and the LLFF dataset~\cite{mildenhall2019local} containing real captures of real-world scenes. The NeRF synthetic dataset is composed of rendering from eight objects with complicated geometry and realistic non-Lambertian materials. The objects are rendered from viewpoints sampled on the upper hemisphere or a full sphere, and each scene has 100 training views and 200 testing views with a spatial resolution of 800$\times$800 pixels. We use the testing views of the eight objects in our experiments. The LLFF dataset contains handheld forward-facing captures of eight complex real-world scenes from the set up in~\cite{mildenhall2020nerf}, and each scene has 20 to 62 images with spatial resolution of 1008 $\times$ 756 pixels. We resize the images to 960 $\times$ 640 pixels for testing following MVSNeRF~\cite{chen2021mvsnerf}. Unless otherwise stated, we use nearest $M=3$ source views as inputs.

\begin{figure}[t]
\begin{center}
\includegraphics[width=0.98\linewidth]{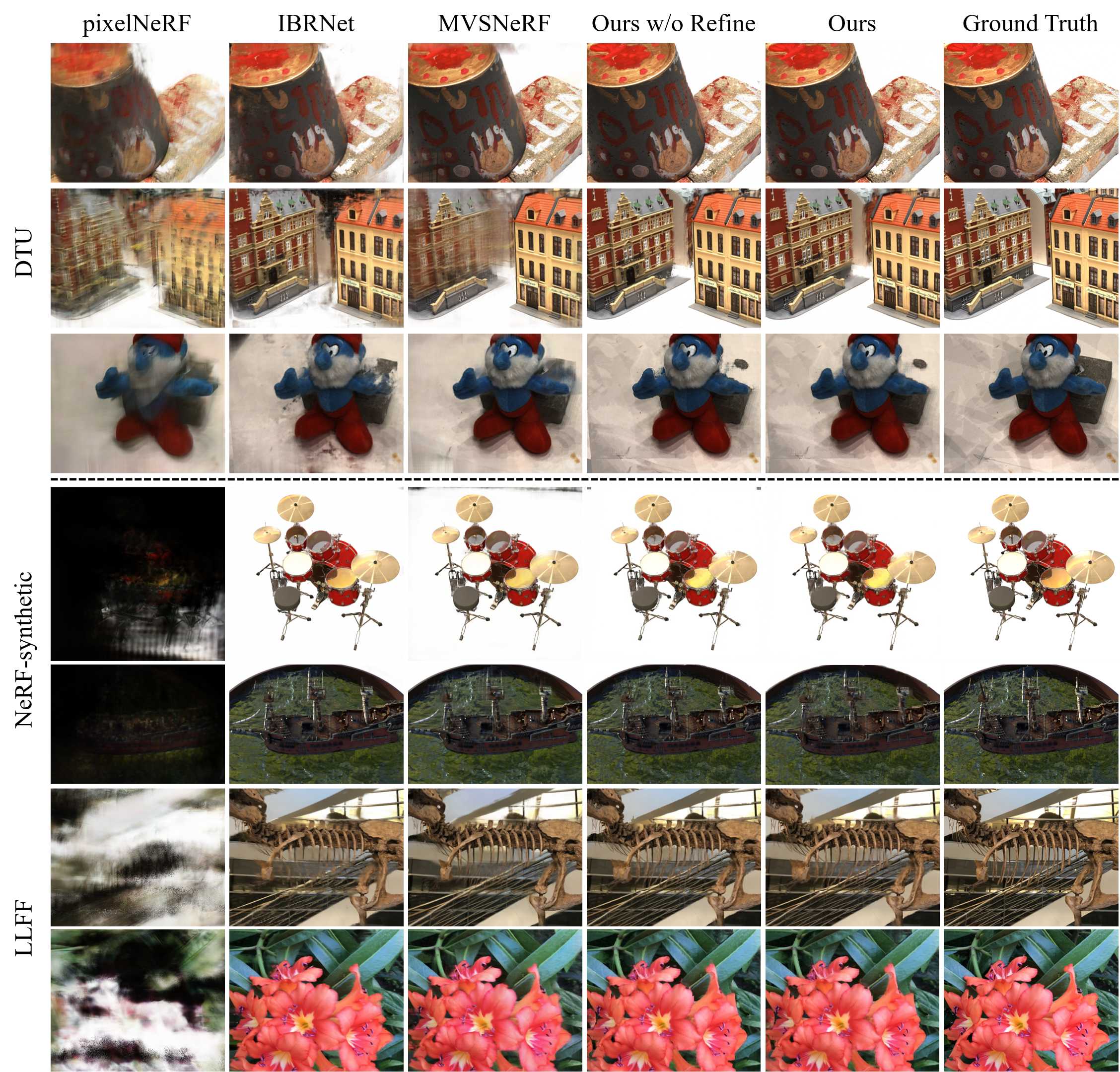}
\end{center}
\vspace{-3mm}
\caption{Qualitative comparison on DTU, NeRF-synthetic and LLFF data. Compared with the state-of-the-art general NVS methods, our ProbNVS achieves photo-realistic novel view synthesis with finer details in terms of geometry and appearance, and also generalizes to new synthetic and real data.}
\vspace{-3mm}
\label{fig:comparison}
\end{figure}

\begin{table}[t]
\begin{center}
\caption{Quantitative results of generalization on NeRF Synthetic and LLFF datasets.}
\label{table:general}
\setlength{\tabcolsep}{1mm}
\begin{threeparttable}
\begin{tabular}{l c c c c c c}
\hline
\multicolumn{1}{l}{\multirow{2}{*}{Method}} & \multicolumn{3}{c}{NeRF Synthetic} & \multicolumn{3}{c}{Real Data (Forward-Facing)}\\ \cmidrule(r){2-4} \cmidrule(r){5-7}
                         & PSNR$\uparrow$ & SSIM$\uparrow$ & LPIPS$\downarrow$ & PSNR$\uparrow$ & SSIM$\uparrow$ & LPIPS$\downarrow$ \\ 
\hline
pixelNeRF~\cite{yu2021pixelnerf}  & 1.4296 & 0.0865 & 0.5683 & 9.1673 & 0.3309 & 0.7054\\
IBRNet\tnote{*}~\cite{wang2021ibrnet}      & 30.4011\tnote{*} & 0.9599\tnote{*} & 0.0678\tnote{*} & 24.4858\tnote{*} & 0.8145\tnote{*} & 0.2080\tnote{*}\\
MVSNeRF~\cite{chen2021mvsnerf}    & 26.0588 & 0.9515 & 0.1039 & 21.7880 & 0.7131 & 0.3179\\
\hline
ProbNVS w/o Refine                     & 25.9989 & 0.9397 &    0.0930 & 21.7323 & 0.7658 & 0.2330\\
ProbNVS                           & \textbf{26.3884} & \textbf{0.9431} & \textbf{0.06360} & \textbf{21.9959} & \textbf{0.7709} & \textbf{0.2327}\\
\hline
\end{tabular}
\begin{tablenotes}
\scriptsize  \item[*]IBRNet is trained on various datasets containing both synthetic and real data, achieving better generalization results. pixelNeRF, MVSNeRF and our ProbNVS are trained only on the DTU dataset.
\end{tablenotes}
\end{threeparttable}
\end{center}
\vspace{-4mm}
\end{table}

\subsection{Results and Comparisons}
\label{sec:comparison}

\noindent\textbf{Comparisons on DTU.}
We compare our ProbNVS with three recent generic novel view synthesis work, pixelNeRF~\cite{yu2021pixelnerf}, IBRNet~\cite{wang2021ibrnet} and MVSNeRF~\cite{chen2021mvsnerf}. We adopt the released code and trained model of each work. In detail, the pixelNeRF and MVSNeRF are trained on DTU similar to ours, and the IBRNet is trained on four different datasets ranging from synthetic to real data with strong generalization ability. We test all methods with $M=3$ nearby source views 
as inputs, and all methods are tested on a single GTX 3090 GPU. We first test all the methods on the test set of DTU dataset~\cite{aanaes2016large}, where we employ the same test setup for each method
and calculate PSNR, SSIM and LPIPS as metrics. We demonstrate the quantitative and qualitative comparison in Tab.~\ref{table:dtu} and Fig.~\ref{fig:comparison}, respectively. In terms of the rendering efficiency, our ProbNVS takes much fewer sampling points for rendering~(eight samples V.S. more than one hundred samples in the other works), and the rendering speed is about 15$\times$, 38$\times$ and 338$\times$ faster than MVSNeRF, IBRNet and pixelNeRF. More importantly, our method is numerically equivalent or even better than existing methods as shown in Tab.~\ref{table:dtu}. Our method can achieve visually high-quality results with much fewer sampling points as shown in Fig.~\ref{fig:comparison}. Such results are attributed to efficient probability-guided sampling, which helps identify the most important set of points for volume rendering. Furthermore, the confidence-aware refinement leverages the uncertainty of the learned depth probability distribution to further improve the rendering of the uncertain, occluded and unreferenced regions.

\noindent\textbf{Generalization.}
For testing the generalization ability of our method, we also show quantitative and qualitative results on other two datasets, the realistic synthetic data from NeRF~\cite{mildenhall2020nerf} and forward-facing real data from LLFF~\cite{mildenhall2019local}. The PSNR, SSIM and LPIPS metrics are reported in Tab.~\ref{table:general}, and the visual comparisons are presented in the lower part of Fig.~\ref{fig:comparison}. Although our method is only trained on the DTU dataset, it generalizes well to new datasets with significantly different data, and shows better numerical and visual results than pixelNeRF and MVSNeRF.
PixelNeRF tends to overfit the DTU dataset and certain coordinate systems, resulting in poor performance. Note that although IBRNet shows the best performance, it is trained on four different datasets with both synthetic and real data, whereas our method is trained only on DTU.

\begin{table}[t]
\begin{center}
\caption{Quantitative ablation study on ProbNVS.}
\label{table:ablation}
\begin{tabular}{l|c c c|c c c}
\hline
Algorithm & PGS & Depth-sup & Refinement & PSNR & SSIM & LPIPS \\
\hline
w/o PGS (single peak) &  & \checkmark &  & 19.0473 & 0.6681 & 0.3960\\
w/o Depth-Sup & \checkmark & & & 21.8337 & 0.6335 & 0.4295\\
Freeze Sampling & \checkmark$\star$ & \checkmark & & 22.5813 & 0.7952 & 0.2804\\
w/o Refine & \checkmark & \checkmark & & 23.8093 & 0.8167 & \underline{0.2622}\\
w Dir-Refine & \checkmark & \checkmark & $\star$ & \underline{24.2247} & \underline{0.8195} & 0.2823 \\
ProbNVS & \checkmark & \checkmark & \checkmark & \textbf{24.6635} & \textbf{0.8316} & \textbf{0.2516} \\
\hline
\end{tabular}
\end{center}
\vspace{-4mm}
\end{table}

\begin{figure}[t]
\begin{center}
\includegraphics[width=0.98\linewidth]{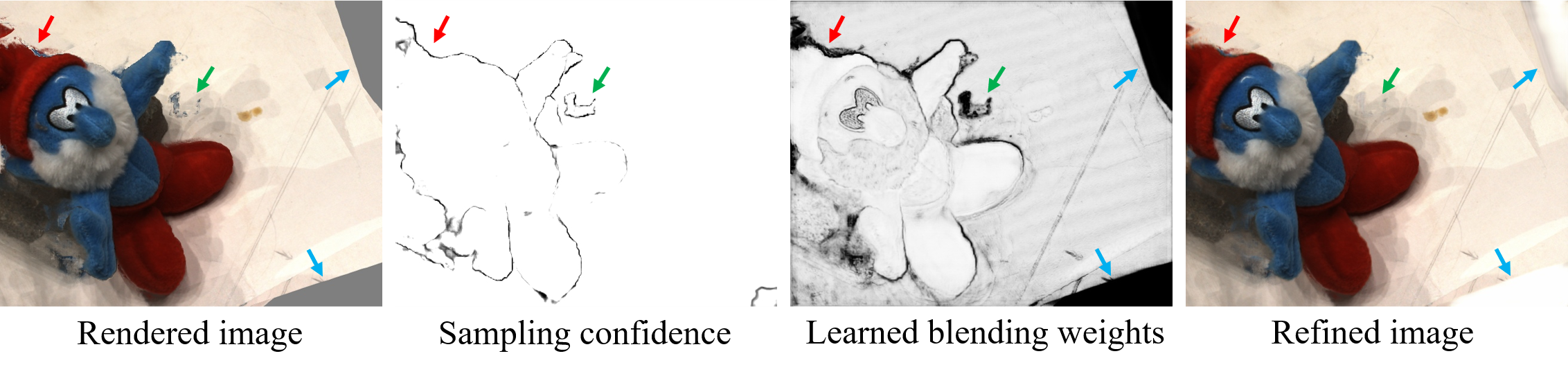}
\end{center}
\vspace{-4mm}
\caption{A case for confidence-aware refinement. The confidence-aware refinement module takes a rendered image and a confidence map calculated from the learned depth probability distributions as inputs, outputs blending weights to generate a refined image. The occluded~(red arrow), uncertain~(green arrow) and unreferenced~(blue arrows) regions can be refined with reasonable colors.}
\label{fig:refine}
\vspace{-4mm}
\end{figure}

\subsection{Ablations and Analysis}
\label{sec:ablation}

\noindent\textbf{Ablation studies.} In this section, we conduct ablation studies to investigate different components of our ProbNVS framework on the test set of DTU, and the metrics are recorded in Tab.~\ref{table:ablation}. Specifically, we first replace our probability-guided sampling~(PGS) with the single peak depth assumption where points are sampled in a single depth interval defined by the depth prediction and its variance, noted as ``w/o PGS (single peak)''. The single peak assumption shows worse results in the test scenes, as depth predictions from only three input views are more likely to fail into wrong locations, leading to incorrect rendering. For ``w/o Depth-Sup'', we remove the depth-supervised pre-training of the depth probability-sampling part, and only use the image loss to train the entire rendering pipeline. As shown in Tab.~\ref{table:ablation}, it is hard for the rendering part to produce good results without proper sampling initialization to give the network a learned depth prediction prior. In contrast, if we freeze the learned sampling and train only the rendering part~(``Freeze sampling''), the results are not good neither, indicating that our framework benefits from both depth-supervised sampling initialization and guidance for correct rendering. Thanks to our training schedule, ``ProbNVS w/o Refine'' achieves better results than previous settings, validating the effectiveness of the entire rendering process. In order to further improve the results of artifacts in uncertain, occluded and unreferenced areas when the input source views are limited~(e.g., $M=3$), we propose a confidence-aware refinement module where the confidence is obtained from the probability-guided sampling module. The final ``ProbNVS'' shows the highest quantitative and visual results, while ``ProbNVS + Dir-Refine'' without confidence input only improves PSNR and SSIM, but fails to improve LPIPS. Please refer to the supplemental material for visual comparisons.

\noindent\textbf{Analysis.} To investigate the performance of our ProbNeRF as a function of the number of rendered samples and input views, we also conduct several experiments and the results are presented in Tab.~\ref{table:num_samples}. On the one hand, we vary the number of points on each ray for rendering as ``1, 2, 4, 8''. The metrics show that more samples lead to better results with a negligible increasing of rendering time. On the other hand, we change the number of the nearest source views for NVS to ``3, 5, 7'' to show the relationship of performance and the number of input viewpoints. As shown in the right part of Tab.~\ref{table:num_samples}, increasing the number of source views provides better results, but slows down rendering.

\begin{table}[t]
\begin{center}
\caption{Analysis on number of samples/views (ProbNVS w/o refine tested on DTU).}
\label{table:num_samples}
\begin{tabular}{l|cccc|ccc}
\hline
\multicolumn{1}{l|}{\multirow{2}{*}{Metric}} & \multicolumn{4}{c|}{Number of samples} & \multicolumn{3}{c}{Number of views} \\
                             & 1   & 2   & 4   & 8  & 3   & 5   & 7\\
\hline
PSNR                    & 21.7675  & 21.7794 & 23.3679 & \textbf{23.8093} & 23.8093 & 24.9202 & \textbf{25.0378} \\
SSIM                    & 0.7935  & 0.7937  & 0.8031 & \textbf{0.8167}  & 0.8167  & 0.8322  & \textbf{0.8336}  \\
LPIPS                   & 0.2681  & 0.2679  & 0.2644 & \textbf{0.2622}  & 0.2622  & 0.2496  & \textbf{0.2469}  \\
\hline
Rendering speed (fps)   & \textbf{6.8117}  & 6.7281 & 6.5596 & 6.5342  & \textbf{6.5342}  & 5.1084  & 4.2328  \\
\hline
\end{tabular}
\end{center}
\vspace{-6mm}
\end{table}

\section{Conclusions}
In this paper, we present a method for novel view synthesis with fast rendering speed, strong generalization capacity and photo-realistic rendering performance. 
We demonstrate that the proposed probability-guided sampling method, combined with the subsequent neural rendering module, greatly speeds up the rendering without losing realistic rendering quality and generalization capacity. 
Moreover, in the case of limited input source views and severe occlusions, the confidence-aware refinement module can optimize uncertain, occluded and unreferenced regions after neural volume rendering. 
Finally, our method also supports different number of input source views and sampling points. 
Regarding to the limitation, the proposed refinement module cannot exactly guarantee cross-view rendering consistency, and the rendering speed still needs to be improved to fulfill higher resolution rendering requirements, which we leave as future work.
To conclude, our ProbNVS contribute a new framework for efficient sampling and high-quality neural volume rendering for generic novel view synthesis tasks.

\clearpage
%
%
\bibliographystyle{splncs04}
\bibliography{egbib}
\end{document}